# Solution of Rectangular Fuzzy Games by Principle of Dominance Using LR-type Trapezoidal Fuzzy Numbers

Arindam Chaudhuri
Lecturer (Mathematics and Computer Science),
Meghnad Saha Institute of Technology,
Nazirabad, Uchchepota, Kolkata, India
Email: arindam_chau@yahoo.co.in

**ABSTRACT**

**Fuzzy Set Theory has been applied in many fields such as Operations Research, Control Theory, and Management Sciences etc. In particular, an application of this theory in Managerial Decision Making Problems has a remarkable significance. In this Paper, we consider a solution of Rectangular Fuzzy game with pay-off as imprecise numbers instead of crisp numbers viz., interval and LR-type Trapezoidal Fuzzy Numbers. The solution of such Fuzzy games with pure strategies by minimax-maximin principle is discussed. The Algebraic Method to solve $2 \times 2$ Fuzzy games without saddle point by using mixed strategies is also illustrated. Here, $m \times n$ pay-off matrix is reduced to $2 \times 2$ pay-off matrix by Dominance Method. This fact is illustrated by means of Numerical Example.**

## 1. INTRODUCTION

The problem of Game Theory [1, 2] is defined as a body of knowledge that deals with making decisions when two or more intelligent and rational opponents are involved under conditions of conflict and competition. In practical life, it is required to take decisions in a competing situation when there are two or more opposite teams with conflicting interests and the outcome is controlled by the decisions of all parties concerned. Such problems occur frequently in Economics, Business Administration, Sociology, Political Science and Military Operations [2]. In all the above problems of different Sciences where the competitive situations are involved, one acts in a rational manner and tries to resolve the conflict of interests in his favour. It is in this context that the Game Theory can prove a useful decision making tool. Instead of making inferences from the past behavior of the opponent, the approach of Game Theory is to seek to determine a rival's most profitable counter strategy to one's own best moves and formulate the appropriate defensive measures. For Example, if two firms are locked up in a war to maintain their market share, then a price-out by the first firm will invite reaction from the second firm in the nature of a price-cut. This will, in turn, affect the sales and profits of the first firm which will again have to develop a counter strategy to meet the challenge from the second firm. The game will thus go on until one of the firms emerges as winner.

The mathematical treatment of the Game Theory was made available in 1944, when John Von Newmann and Oscar Morgenstern published the famous article *Theory of Games and Economic Behavior* [2]. The Von Newmann's approach to solve the Game Theory problems was based on the principle of *best out of the worst* i.e., he utilized the idea of minimization of the maximum losses. Most of the competitive problems can be handled by this principle. However, in real life situations, the information available is of imprecise nature and there is an inherent degree of vagueness or uncertainty present in the system under consideration. In order to tackle this uncertainty the concept of Fuzzy Sets can be used as an important decision making tool. Imprecision here is meant in the sense of vagueness rather than the lack of knowledge about the parameters present in the system. The Fuzzy Set Theory thus provides a strict mathematical framework in which vague conceptual phenomena can be precisely and rigorously studied.

In this work, we have concentrated on the solution of Rectangular Games by Principle of Dominance Using LR-type Trapezoidal Fuzzy Numbers. The LR-type Trapezoidal Fuzzy Numbers are defined by the Trapezoidal Membership Functions. They are characterized by their simple formulations and computational efficiency and thus have been used extensively to solve different problems in Engineering and Management. The solution of Fuzzy games with Pay-off as imprecise number is generally given by the minimax-maximin Principle. In the past, work has also been done on Algebraic Method of solving $m \times n$ Rectangular Fuzzy games with Pay-off as Interval Numbers having no Saddle Point. Determination of $2 \times 2$ Fuzzy games from a Rectangular $m \times n$ Fuzzy game without Saddle Point is a fundamental problem of Fuzzy Game Theory. In Classical Game Theory the Pay-off is a Crisp Number. Practically, it may happen that Pay-off is not necessarily a fixed real number. Here, the Pay-off is considered as LR-type Trapezoidal Fuzzy Number and the $m \times n$ matrix is reduced to $2 \times 2$ matrix. The Organization of this Paper is as follows. In section 2, we define the Trapezoidal Membership Function. In the next section, we consider the concept of the Interval Numbers. In the section 4, we give the basic definitions of the Two Person Zero Sum Games and the Pay-off Matrix. The section 5 discusses Solution of $2 \times 2$ Games with mixed strategies. In the next section, we discuss about Games having no Saddle Point. The section 7 illustrates the Concept of Dominance. This is followed by a Simulation Example. Finally, in section 8 conclusions are given.

## 2. TRAPEZOIDAL MEMBERSHIP FUNCTION

A trapezoidal membership function is defined by four parameters viz., $a, b, c, d$ as follows [3]:

$$trapezoid(x; a,b,c,d) = \begin{cases} 0, x \leq a \\ \frac{x-a}{b-a}, a \leq x \leq b \\ 1, b \leq x \leq c \\ \frac{d-x}{d-c}, c \leq x \leq d \\ 0, d \leq x \end{cases} ; x \in \Re$$

## 3. INTERVAL NUMBERS

An interval number [4] is defined as $I = [X_L, X_R] = \{x : X_L \leq x \leq X_R, x \in \Re\}$. Another way of representing an interval number in terms of midpoint is, $I = <m(i), w(i)>$ where, $m(i) =$ midpoint of $I = (X_L + X_R)/2$ and $w(i) =$ half width of $I = (X_L - X_R)/2$. Addition of two interval numbers $I = [X_L, X_R]$ and $J = [Y_L, Y_R]$ is $I + J = [X_L + Y_L, X_R + Y_R]$. Using mean width notations, if $I = <m_1, w_1>$ and $J = <m_2, w_2>$ then $I + J = <m_1 + m_2, w_1 + w_2>$. Similarly, the other binary operations on interval numbers are defined [4].

## 3.1 ORDERED RELATION AMONG INTERVALS

If $I = [a,b]$, $J = [c,d]$ then $[a,b] < [c,d]$, iff $b < c$ and is denoted by $I < J$. $I$ is contained in $J$ iff $a \geq c, b \leq d$ and this is denoted by $I \leq J$.

*Definition* 1: The *dominated index* $(DI)$ to proposition $I$ is dominated over $J$ as $I(I < J) = (m_2 - m_1)/(w_1 + w_2)$.

Using $DI$ the following ranking order is defined.

*Definition* 2: If $DI(I < J) \geq 1$, then $I$ is said to be totally dominating over $J$ in the sense of minimization and $J$ is said to be totally dominating over $J$ in the sense of minimization. This is denoted by $I < J$.

*Definition* 3: If $0 < DI(I < J) < 1$, then $I$ is said to be partially dominating over $J$ in the sense of minimization and $J$ is said to be partially dominating over $J$ in the sense of minimization. This is denoted by $I < J$.

When $DI(I < J) = 0$, then $m_1 = m_2$, it may be emphasized on the width of interval numbers $I$ and J. If $w_1 < w_2$, then left end point of $I$ is less than that of $J$ and there is a chance that on finding a minimum distance, the distance may be on $I$. But at the same time, since the right end point of $I$ is greater than that of $J$, if one prefers $I$ over $J$ in minimization then in worst case, he may be looser than one who prefer $J$ over $I$.

*Numerical Example*:

$I = [110, 120] = <115, 5>$, $J = [150, 155]$
$= <152.5, 2.5>$, $DI(I < J) = (152.5 - 115)/2 > 1$. So in minimization $I$ is totally dominating over $J$.

*Definition* 4: The *dominated index* ($DI$) of proposition $A = (\alpha, a, \beta)$ is dominated over $B = (\gamma, b, \delta)$ is given by $DI(A < B) = (b - a)/(\beta + \gamma)$. Using $DI$ index the following ranking order is defined.

*Definition* 5: If $DI(A < B) \geq 1$, $A$ is said to be totally dominating over $B$ in the sense of minimization and $B$ is said to be totally dominating over $A$ in the sense of maximization, it is also denoted by $A < B$.

*Definition* 6: If $0 < DI(A < B) < 1$, then $A$ is said to be partially dominating over $B$ in the sense of minimization and $B$ is said to be partially dominating over $A$ in the sense of maximization, it is also denoted by $A < B$.

*Lemma1*: If $DI(A < B) = 0$ then $A$ and $B$ are said to be non comparable and is denoted by $A \neq B$. In this case $A$ is preferred over $B$ if (a) $\alpha = \gamma$ and $\beta < \delta$ or (b) $\alpha > \gamma$ and $\beta = \delta$, otherwise a pessimistic decision maker would prefer the number with smaller length of support whereas an optimistic decision maker would do the converse.

*Numerical Example*:

(1) If $A = <6, 0.2>$, $B = <1, 0.2>$ then $DI(A < B) = (6 - 1)/(0.2 + 0.2) > 1$. Thus, $A$ is totally dominating over $B$ in the sense of minimization and $B$ is said to be totally dominating over $A$ in the sense of maximization.

(2) If $A = <0.4, 0.5>$, $B = <0.3, 0.5>$ then
$DI(A < B) = (0.4 - 0.3)/(0.5 + 0.5) = 0.1$
$DI(A < B) = (0.4 - 0.3)/(0.5 + 0.5) = 0.1$. So, $A$ is said to be partially dominating over $B$.

## 4. TWO PERSON ZERO SUM GAMES AND PAY-OFF MATRIX

In this section we give some basic definitions of the Two Person Zero Sum Games and pay-off matrix. These concepts form the basic building blocks of Game Theory.

### 4.1 TWO PERSON ZERO SUM GAMES

A game of two persons in which gains of one player are losses of other player is called a *two person zero sum game*, i.e., in two person zero sum the algebraic sum of gains to both players after a play is bound to be zero. Games relating to pure strategies taken

by players are considered here based on two assumptions [1, 2]:

1. Player $A$ is in better position and is called maximization player (or row player) and player $B$ is called minimizing player (or column player).

2. Total gain of one player is exactly equal to total loss of other player. In general, if player $A$ takes $m$ pure strategies and $B$ takes $n$ pure strategies, then the game is called *two person zero sum game* or $m \times n$ *rectangular game*.

**4.2 PAY-OFF MATRIX**

*Two person zero sum games* are known as *rectangular games* since they are represented by rectangular pay-off matrix. A pay-off matrix is always written for maximizing player. Considering the general $m \times n$ rectangular game, the pay-off matrix of $A$ with $m$ pure strategies $A_1, \ldots, A_m$ and $B$ with $n$ pure strategies $B_1, \ldots, B_n$ is given by [3],

$$\begin{bmatrix} <a_{11},b_{11}> \ldots <a_{1n},b_{1n}> \\ \ldots \\ \ldots \\ \ldots \\ \ldots \\ \ldots \\ <a_{m1},b_{m1}> \ldots <a_{mn},b_{mn}> \end{bmatrix}$$

The elements $<a_{ij},b_{ij}>$ are LR-type trapezoidal fuzzy numbers and for crisp game they may be positive, negative or zero. When player $A$ chooses strategy $A_i$ and player $B$ selects $B_j$, it results in pay-off of LR-type trapezoidal fuzzy number $<a_{ij},b_{ij}>$ to player $A$.

**5. SOLUTION OF $2 \times 2$ GAMES WITH MIXED STRATEGIES**

Consider the fuzzy game [3, 5] of players A (strategies represented horizontally) and B (strategies represented vertically) whose pay-off is given by following matrix and for which there is no saddle point.

$$\begin{bmatrix} <a_{11},b_{11}> & <a_{12},b_{12}> \\ <a_{21},b_{21}> & <a_{22},b_{22}> \end{bmatrix}$$

where, pay-off $<a_{ij},b_{ij}>$ are symmetric LR-type trapezoidal fuzzy numbers such that $b_{ij} = a_{ij} + \alpha$. If $x_i$ and $y_j$ be the probabilities by which $A$ chooses $i^{th}$ strategy and $B$ chooses $j^{th}$ strategy then:

$$x_1 = (a_{22} - a_{21})/(a_{11} + a_{22} - a_{12} - a_{21});$$
$$y_1 = (a_{22} - a_{21})/(a_{11} + a_{22} - a_{12} - a_{21});$$
$$x_2 = (a_{11} - a_{12})/(a_{11} + a_{22} - a_{12} - a_{21});$$
$$y_2 = (a_{11} - a_{12})/(a_{11} + a_{22} - a_{12} - a_{21});$$

which are crisp numbers and value of the game can be easily computed as $V = <a,b>$; where, $a$ and $b$ are left and right spreads of LR-type trapezoidal fuzzy numbers given by:

$$a = (a_{11}a_{22} - a_{12}a_{21})/(a_{11} + a_{22} - a_{12} - a_{21});$$
$$b = (b_{11}b_{22} - b_{12}b_{21})/(b_{11} + b_{22} - b_{12} - b_{21})$$

**6. GAMES WITH NO SADDLE POINT**

The simplest case is a $2 \times 2$ Fuzzy game with no saddle point. Here, we consider a $m \times n$ Fuzzy game. Now we discuss a particular method. In this method, the pay-off can be reduced to $2 \times 2$ games so that it can be solved by using the Fuzzy game method. The method of reduction of the pay-off matrix by this process is called the Dominance property [1, 2] of the rows and columns of the pay-off matrix.

# 7. CONCEPT OF DOMINANCE

If one pure strategy of a player is better for him or as good as another, for all possible pure strategies of opponent then first is said to dominate the second [1, 2]. The dominated strategy can simply be discarded from pay-off matrix since it has no value. When this is done, optimal strategies for the reduced matrix are also optimal for the original matrix with zero probability for discarded strategies. When there is no saddle point in pay-off matrix, then size of the game can be reduced by dominance, before the problem is solved.

*Definition* 7: If all elements of the $i^{th}$ row of pay-off matrix of a $m \times n$ rectangular game are dominating over $r^{th}$ row in the sense of maximization, $r^{th}$ row is discarded and deletion of $r^{th}$ row from matrix does not change the set of optimal strategies of maximizing player.

*Numerical Example*: Consider the fuzzy game of two players A (strategies represented horizontally) and B (strategies represented vertically) with the following pay-off matrix. Player A is maximizing player and player B is minimizing player.

$$\begin{bmatrix} <1,0.2> & <7,0.3> & <2,0.1> \\ <6,0.2> & <2,0.1> & <7,0.3> \\ <0,0.2> & <1,0.2> & <6,0.2> \end{bmatrix}$$

$$DI(A_{31} < A_{21}) = (6-0)/(0.2+0.2) > 1$$
$$DI(A_{32} < A_{22}) = (2-1)/(0.4+0.2) > 1$$
$$DI(A_{31} < A_{21}) = (7-6)/(0.3+0.2) > 1$$

Thus $A_2$ is dominating over $A_3$ in the sense of maximization and row $A_3$ is deleted. The reduced matrix is given by,

$$\begin{bmatrix} <1,0.2> & <7,0.3> & <2,0.1> \\ <6,0.2> & <2,0.1> & <7,0.3> \end{bmatrix}$$

*Definition* 8: If all elements of $j^{th}$ column are dominating over $s^{th}$ column in the sense of minimization the $s^{th}$ column is deleted and the deletion of $s^{th}$ column from the matrix does not change the set of optimal strategies of minimizing player.

*Numerical Example*:

Considering the above pay-off matrix,

$$DI(B_{11} < B_{13}) = (2-1)/(0.1+0.3) > 1$$
$$DI(B_{21} < B_{23}) = (7-6)/(0.2+0.3) > 1$$

Here $B_1$ is totally dominating over $B_3$ in the sense of minimization and the resultant pay-off matrix is given by,

$$\begin{bmatrix} <1,0.2> & <7,0.3> \\ <6,0.2> & <2,0.1> \end{bmatrix}$$

*Definition* 9: If the linear combination of $p^{th}$ and $q^{th}$ rows dominates all elements of the $s^{th}$ row in the sense of minimization, $s^{th}$ row is discarded and the deletion of $s^{th}$ row from matrix does not change the set of optimal strategies of maximizing player.

*Numerical Example*: Considering a particular pay-off matrix of two players A (strategies represented horizontally) and B (strategies represented vertically) as follows:

$$\begin{bmatrix} <1,0.4> & <2,0.1> & <-1,0.1> \\ <3,0.5> & <1,0.3> & <2,0.2> \\ <-1,0.2> & <3,0.4> & <2,0.4> \end{bmatrix}$$

The convex combination of second and third row gives $A_4 = \beta A_3 + (1-\beta)A_3; 0 \leq \beta \leq 1$.

Taking $\beta = 0.5$ the elements of $A_4$ are $<1,0.35>, <2,0.35>$ and $<2,0.30>$ Now $A_4$ is dominating over $A_1$ in the sense of maximization and row $A_1$ is discarded such that the resulting pay-off matrix is given by,

$$\begin{bmatrix} <3,0.5> & <1,0.3> & <2,0.2> \\ <-1,0.2> & <3,0.4> & <2,0.4> \end{bmatrix}$$

*Definition* 10: If $j^{th}$ column is dominated by the convex combination of $m^{th}$ and $n^{th}$ column, $j^{th}$ column is discarded in sense of minimization and deletion of $j^{th}$ column from matrix does not change the set of optimal strategies of the minimizing player.

*Numerical Example*:

Considering above pay-off matrix, the convex combination of $B_1$ and $B_2$ i.e., $B_4 = \alpha B_1 + (1-\alpha) B_2; 0 \le \alpha \le 1$.

Taking $\alpha = 0.5$ elements of $B_4$ are,

$$\begin{bmatrix} <2,0.4> \\ <1,0.3> \end{bmatrix}$$

Now $B_4$ is totally dominating over $B_3$ and thus the third column is removed such that the resulting matrix is given by,

$$\begin{bmatrix} <3,0.5> & <3,0.4> \\ <-1,0.2> & <1,0.3> \end{bmatrix}$$

*Definition* 11: When there is no saddle point and no course of action dominates any other the values for different $2 \times 2$ sub games are computed. As $A$ is maximizing player he will definitely select that pair strategies which will give the best value of $2 \times 2$ sub games and the corresponding sub matrix provides optimal solution. Similarly, $B$ is minimizing player he will definitely select that pair of courses, which will give the least value of $2 \times 2$ sub games, the corresponding sub matrix will provide optimal solution to the fuzzy problem.

*Numerical Example*:

Consider the fuzzy game whose pay-off matrix is given by,

$$\begin{bmatrix} <19,0.2> & <15,0.4> & <16,0.1> \\ <0,0.2> & <20,0.4> & <5,0.4> \end{bmatrix}$$

There is no saddle point and no course of action dominates any other. The values $V_1$, $V_2$, $V_3$ are computed from the following three $2 \times 2$ sub games as obtained from given matrix.

*Sub game 1*: $\begin{bmatrix} <19,0.2> & <15,0.4> \\ <0,0.2> & <20,0.4> \end{bmatrix}$;

The corresponding value $V_1 = <\frac{95}{6}, \frac{7}{5}>$.

*Sub game 2*: $\begin{bmatrix} <19,0.2> & <16,0.1> \\ <0,0.2> & <5,0.4> \end{bmatrix}$;

The corresponding value $V_2 = <16,0.1>$.

*Sub game 3*: $\begin{bmatrix} <15,0.4> & <16,0.1> \\ <20,0.4> & <5,0.4> \end{bmatrix}$;

The corresponding value $V_3 = <\frac{245}{16}, \frac{11}{5}>$.

Here, $\min\{V_1, V_2, V_3\} = V_3$ such the resulting pay-off matrix is:

$$\begin{bmatrix} <15,0.4> & <16,0.1> \\ <20,0.4> & <5,0.4> \end{bmatrix}$$

## 8. NUMERICAL SIMULATION

To illustrate dominance method, we consider the following pay-off matrix:

$$\begin{bmatrix} <8,0.3> & <15,0.4> & <-4,0.1> & <-2,0.4> \\ <19,0.1> & <15,0.5> & <17,0.4> & <16,0.1> \\ <0,0.3> & <20,0.2> & <15,0.5> & <5,0.4> \end{bmatrix}$$

All $DI(A_1 < A_2) \geq 1$, so $A_2$ is totally dominating over $A_1$ in the sense of minimization and row $A_1$ is deleted, such that the resulting pay-off matrix is:

$$\begin{bmatrix} <19,0.1> & <15,0.5> & <17,0.4> & <16,0.1> \\ <0.0.3> & <20,0.2> & <15,0.5> & <5,0.4> \end{bmatrix}$$

Again all $DI(B_4 < B_3) \geq 1$, so $B_4$ is totally dominating over $B_3$ in the sense of minimization and the column $B_3$ is deleted such that the resulting pay-off matrix is:

$$\begin{bmatrix} <19,0.1> & <15,0.5> & <16,0.1> \\ <0.0.3> & <20,0.2> & <5,0.4> \end{bmatrix}$$

This is no course of action which dominates any other and there is no saddle point. The values for different $2 \times 2$ pair of strategies are computed. Since, $B$ is minimizing player the minimum value is considered and corresponding pay-off matrix provides optimal solution to fuzzy problem. The previous least value of $\{V_1, V_2, V_3\}$ is $V_3$ so optimal strategies of $A$ are $(A_2, A_3)$ and of $B$ are $(B_2, B_3)$. The final pay-off matrix is given by,

$$\begin{bmatrix} <15,0.5> & <16,0.1> \\ <20,0.2> & <5,0.4> \end{bmatrix}$$

The probabilities are $x_1 = 0, x_2 = \frac{15}{16}, x_3 = \frac{1}{16}, y_1 = 0, y_2 = \frac{1}{16}, y_3 = 0, y_4 = \frac{15}{16}$

and value of the game is $V = <\frac{245}{16}, \frac{11}{5}>$. Hence, optimal solution of the complete game is $(0, x_2, x_3)$ for $A$ and $(0, y_2, 0, y_4)$ for $B$; the value of game being $V$.

## 9. CONCLUSION

We considered the solution of Rectangular Fuzzy games using LR-type Trapezoidal Fuzzy Numbers. Here pay-off is considered as imprecise numbers instead of crisp numbers which takes care of the uncertainty and vagueness inherent in such problems. LR-type Trapezoidal Fuzzy Numbers are used because of their simplicity and computational efficiency. We discuss solution of Fuzzy games with pure strategies by minimax-maximin principle and also Algebraic Method to solve $2 \times 2$ Fuzzy games without saddle point by using mixed strategies. The Concept of Dominance Method is also illustrated. LR-type Trapezoidal Fuzzy Numbers generates optimal solutions which are feasible in nature and also takes care of the impreciseness aspect.